\documentclass[sigconf]{acmart}

\usepackage[utf8]{inputenc}
\usepackage[OT4,T1]{fontenc}

\usepackage{balance}

\usepackage{graphicx}

\usepackage{algorithmic}
\usepackage{algorithm}
\usepackage{listings}
\usepackage[OT4,T1]{fontenc}
\usepackage{amsmath}

\title[Cyberbullying Detection -- Technical Report 2/2018 CS AGH]{Cyberbullying Detection -- Technical Report 2/2018 \\
Department of Computer Science \\
AGH University of Science and Technology}

\author{Michał Ptaszyński}
\affiliation{%
  \institution{Department of Computer Science,\\ Kitami Institute of Technology}
}
\email{ptaszynski@cs.kitami-it.ac.jp}
\author{Gniewosz Leliwa}
\affiliation{%
  \institution{Samurai Labs/Fido Voice}
}
\email{gniewosz.leliwa@fido.ai}
\author{Mateusz Piech}
\affiliation{%
  \institution{Department of Computer Science,\\ AGH University of Science and Technology}
}
\email{mpiech@agh.edu.pl}
\author{Aleksander Smywiń\-ski-Pohl}
\affiliation{%
  \institution{Department of Computer Science,\\ AGH University of Science and Technology}
}
\email{apohllo@agh.edu.pl}

\begin{abstract}
  The research described in this paper concerns automatic cyberbullying detection in social media.
  There are two goals to achieve: building a gold standard cyberbullying detection dataset and measuring the performance
  of the Samurai cyberbullying detection system. The Formspring dataset provided in a Kaggle competition was
  re-annotated as a part of the research. The annotation procedure is described in detail and, unlike many other recent
  data annotation initiatives,
  does not use Mechanical Turk for finding people willing to perform the annotation. The new annotation compared to the
  old one seems to be more coherent since all tested cyberbullying detection system performed better on the former.

  The performance of the Samurai system is compared with 5 commercial systems and one well-known machine learning
  algorithm, used for classifying textual content, namely Fasttext. It turns out that Samurai scores the best
  in all measures (accuracy, precision and recall), while Fasttext is the second-best performing algorithm.
\end{abstract}

\begin{document}
\maketitle              

\section{Introduction}

Cyberbullying is a phenomenon observed in many Internet services, where people, especially young users, share their
thoughts regarding their personal interests, life and problems. Some of the users use methods such as harassment,
threat, intimidation and mocking in order to make others feel worse, to undermine their self-esteem and to discourage
them from posting questions, messages or personal images. Anonymity is one of the factors which makes this activity
particularly attractive to bullies since they do not face social ostracism or other negative consequences connected with
improper behavior. The problem has been growing since the outset of social networks. The most striking episode occurred
when the Formspring social network was closed, probably due to the suicides connected with the messages containing
cyberbullying, exchanged via that website. It is believed that the feature of the service allowing people who know each
other outside the Internet to message themselves anonymously played an important role.

As such cyberbullying is a very important problem, which has its roots in technology. But technology can also help to
reduce or even eradicate it. Development of Natural Language Processing (NLP) algorithms aimed at the detection of
cyberbullying is one of the ways to achieve that goal. Such algorithms require annotated data at least to measure their
performance. Moreover, the most popular Machine Learning (ML) algorithms, Deep Neural Networks (DNN) in particular,
require large annotated corpora in order to obtain high quality classification results. Unfortunately, there are only a
few publicly available dataset for cyberbullying detection:

\begin{itemize}
  \item Kaggle Formspring data for Cyberbullying
    Detection\footnote{\url{https://www.kaggle.com/swetaagrawal/formspring-data-for-cyberbullying-detection}}
  \item MySpace Group Data Labeled for Cyberbullying\footnote{\url{http://www.chatcoder.com/Data/BayzickBullyingData.rar}}
\end{itemize}

The Kaggle Formspring dataset was originally annotated using the Mechanical Turk service. The methodology behind the annotation
process was pretty simplistic \cite{reynolds2011using}. Namely, a post was marked as containing cyberbullying if two of the annotators
indicated that it contains that phenomenon. Moreover, the annotators did not have any training towards the detection of
cyberbullying. As a result, the annotation has a moderate quality (which we discuss in Section \ref{sec:limitations}). Since the dataset
and its annotation are crucial for the task, we have decided to use the same dataset, but provide a new annotation,
obtained by a well designed process. The new annotation is the primary outcome of our research.

We also want to assess the performance of one of the systems for cyberbullying detection -- namely the Samurai developed
by Samurai Labs (a new brand of Fido Voice). The Samurai's authors point out few key features of their system that
differentiates it from the other state-of-the-art approaches:

\begin{itemize}
  \item The underlying philosophy of dividing the big problem of detecting online violence into a number of smaller
    problems related to certain online violence phenomena, allowing quick adjustment of the system to the clients'
    needs, based on general guidelines how the system should react on certain types of phenomena (``onboarding process'').
  \item The technological ability to enhance the learning capabilities of DNN with a symbolic governance relying on the
    grammar structure and experts’ knowledge, making it much more independent from labeled datasets and allowing to
    overcome many other deep learning's limitations \cite{marcus2018deep}.
  \item The immunity for adversarial attacks \cite{hosseini2017deceiving}.
\end{itemize}

Disclaimer: Samurai Labs has sponsored the annotation initiative. The creators of Samurai did not have access to the new
annotation, until the final testing of the system was performed. In order to keep the highest scientific standards
of independence and objectivity, the process was mediated by the Department of Computer Science of AGH University of
Science and Technology, who performed the final testing of all systems and algorithms versus the old and the new annotation.

\section{Dataset}
\label{sec:dataset}

Formspring data for Cyberbullying Detection (a large unlabeled Formspring dataset, from a Summer 2010 crawl
\cite{reynolds2011using}), available on Kaggle and prepared by Kelly Reynolds was chosen as a test dataset. The main reasons for choosing
this dataset were:
\begin{itemize}
  \item Its comparatively large size: 12772 data samples.
  \item  The fact that it is fairly well-known dataset with an initial release in October, 2016, and the last update in
    January, 2017.
  \item The fact that the number of sentences with bully contents reflects the real-world proportions of cyberbullying
    and no-cyberbullying content (more than 84\% of samples were labeled as ``no cyberbullying'').
  \item  The option of anonymity that encourages cyberbullying and other harmful behaviors (Formspring allowed users to
    post questions anonymously to any other user's page).
  \item  The controversies around Formspring related to harassment and cyberbullying that eventually led to the death of
    few teenagers in 2011 and shutdown of the service in 2013 \footnote{https://en.wikipedia.org/wiki/Spring.me}.
\end{itemize}

          Detailed statistics of the dataset are
presented in Section \ref{sec:stats}.

\subsection{Limitations of the Present Annotation}
\label{sec:limitations}

A preliminary analysis of the annotation quality demonstrates numerous shortcomings that put in question its usage in a
testing process. Each sample was labeled by three Amazon's Mechanical Turk workers with ``yes'' and ``no'' answers for a
question if it contains cyberbullying. The cyberbullying was also tagged for severity from 0
(no bullying) to 10. A post was considered harmful if at least two out of three annotators answered ``yes'' for the
primary question.  As a result 802 samples out of 12772 (6.3\%) were classified as ``cyberbullying''.

In the description of the annotation process there was no information about annotators' competence and there were too
many missed cases of cyberbullying as well as many cases of non-bullying content that were incorrectly labeled as
cyberbullying.

An analysis of the annotated samples showed that at least 2.5\% of posts classified as ``no cyberbullying'' should be labeled as
``cyberbullying'' due to their possible harmful impact. This is a noticeable amount compared to the percentage of the
samples labeled as ``cyberbullying''. Few striking examples of the missed cases are given in Appendix B%
.

Similarly, about 15-20\% of the samples labeled as ``cyberbullying'' could be labeled as ``no cyberbullying'' due to an
infinitesimal harmfulness or even obvious annotators' errors. Few examples of the cases incorrectly labeled as
``cyberbullying'' are given in Appendix B%
.

Therefore, we decided to re-annotate the whole Formspring dataset. The annotators' recruitment process and the actual
re-annotation process are described in details in the following section. The annotation instructions that the annotators
were provided with is available in Appendix A.

\subsection{Annotators}

The task of Cyberbullying Detection (CB-D) is specific in the sense that it requires highly trained data annotators with
sufficient background for high quality annotations. Differently to well known tasks, such as traditional sentiment
analysis, annotators employed in CB-D should either be experienced in Internet Patrol activities (patrolling the
Internet in the search for harmful contents), or should have a sufficient specialist knowledge in psychology,
psychiatry, or related fields.

We made an open call for data annotators within graduate students of psychology, with a condition of at least
near-native English proficiency (language of initial data samples). In specific situations we also allowed
undergraduates, but only with very high achievements.

Sixteen (16) initial candidates responded to our call. The candidates were given an initial test to eliminate possible
outliers and low performance annotators. In the initial test the candidates were given 30 random samples to annotate
with already prepared gold standard answers. The top eight candidates were retained, while the low performance half was
rejected.  Their scores for the initial test are given in Table \ref{tab:annotators}.

\begin{table}
  \begin{center}
    \caption{The number of correctly annotated samples (out of 30) for the people that performed the annotation.}
    \label{tab:annotators}
    \begin{tabular}{|l|r|r|}
      \hline
      \textbf{ID} & \textbf{Correct Labels} & \textbf{Percentage}\\
      \hline
      \#01 & 24 & 80.0\% \\
      \hline
      \#02 & 24 & 80.0\% \\
      \hline
      \#03 & 23 & 76.7\% \\
      \hline
      \#04 & 22 & 73.3\% \\
      \hline
      \#05 & 22 & 73.3\% \\
      \hline
      \#06 & 21 & 70.0\% \\
      \hline
      \#07 & 21 & 70.0\% \\
      \hline
      \#08 & 20 & 66.7\% \\
      \hline
    \end{tabular}
  \end{center}
\end{table}

\subsubsection{I stage of annotation}

Each annotator annotated a large number of samples. The whole annotation process started on May 25th, 2018 and ended on
June 30th, 2018, and was divided into 2 stages. At the first stage, annotators were provided with smaller portions of
the total 12772 samples and each time a deadline for annotation of the given portions was set (usually 6-7 days for
labeling a portion of 2400 samples). The average time for annotating a single sample was 30 seconds. Annotators were
allowed to annotate the given portion in elastic work time, according to their preferences and availability. After each
portion, annotators were asked if they were willing to continue the labeling process with the next portion. They
proceeded only after the agreement. No one resigned during the annotation process.
The assignments of the parts to the annotators in the first stage are given in Table \ref{tab:assignments}.

\begin{table}
  \begin{center}
    \caption{The assignment of samples to individual annotators in the first stage of annotation.}
    \label{tab:assignments}
    \begin{tabular}{|r|l|l|l|}
      \hline
      \multicolumn{1}{|l|}{\textbf{Portion}} &
      \textbf{1$^{st}$ Annot.} &
      \textbf{2$^{nd}$ Annot.} &
      \textbf{3$^{rd}$ Annot.} \\
      \hline
      1-2400 &
      \#08 &
      \#01 &
      \#04 \\
      \hline
      2401-4800 &
      \#04 &
      \#03 &
      \#01 \\
      \hline
      4801-7200 &
      \#01 &
      \#04 &
      \#08 \\
      \hline
      7201-9600 &
      \#03 &
      \#08 &
      \#02 \\
      \hline
      9601-11185 &
      \#06 &
      \#05 &
      \#03 \\
      \hline
      11186-12772 &
      \#07 &
      \#02 &
      \#06 \\
      \hline
    \end{tabular}
  \end{center}
\end{table}

After the first stage of annotations there were 11320 samples (88.63\% of all data) annotated unequivocally by all three
annotators handling each sample. Within this data there were 11007 (86.18\%) samples annotated as
\textit{non-cyberbullying} /
\textit{non-harmful}, 285 (2.23\%) annotated as \textit{cyberbullying / harmful}, and 28 samples annotated as
\textit{uncertain / I don't know}. The
28 samples which all three annotators were not able to annotate were forwarded to a cyberbullying expert for final
decision. Except the unequivocal samples, there were 1452 samples (11.37\%) annotated with some level of disagreement.
This ambiguous data was used in the second stage of the annotation.

\subsubsection{II stage of annotation}

Before the second stage, the preliminary data analysis was performed. At this stage, all of the 1452 ambiguous samples
were distributed between 6 annotators. The parts contained from 467 to 942 samples that corresponded to the primary
division of the data set (e.g. 119 ambiguous samples in the first portion of 1-2400 samples). A deadline for this stage
was set to 4 days. The assignments of the parts to the annotators in the second stage are given in Table
\ref{tab:assignments2}.

\begin{table}
  \begin{center}
    \caption{The assignment of samples to individual annotators in the second stage of annotation.}
    \label{tab:assignments2}
    \begin{tabular}{|r|l|l|l|}
      \hline
      \multicolumn{1}{|l|}{\textbf{Portion}} &
      \textbf{1$^{st}$ Annot.} &
      \textbf{2$^{nd}$ Annot.} &
      \textbf{3$^{rd}$ Annot.} \\
      \hline
      119 from 1-2400 &
      \#03 &
      \#02 &
      \#07 \\
      \hline
      487 from 2401-4800 &
      \#08 &
      \#02 &
      \#07 \\
      \hline
      127 from 4801-7200 &
      \#03 &
      \#02 &
      \#07 \\
      \hline
      209 from 7201-9600 &
      \#04 &
      \#01 &
      \#07 \\
      \hline
      289 from 9601-11185 &
      \#04 &
      \#01 &
      \#08 \\
      \hline
      221 from 11186-12772 &
      \#04 &
      \#01 &
      \#03 \\
      \hline
    \end{tabular}
  \end{center}
\end{table}

In the whole annotation process (both stages), Annotator \#01 to Annotator \#08 annotated in total: 7919, 4720, 6852,
7919, 1585, 3172, 2529, 7976 samples, respectively. This is 42672 samples in total and 5334 samples on average per
annotator.

\subsection{Annotation Guidelines}

Each annotator was provided with a PDF file with the annotation instruction at the beginning of the recruitment process.
The exact instruction as it was delivered to the annotators is presented in the Appendix A.

The main task of an annotator was to label each sample with one of three possible labels:
\begin{itemize}
  \item[0] – text certainly does not contain online violence;
  \item[1] – text certainly contains online violence;
  \item[2] – uncertain case.
\end{itemize}

In both stages, the annotators were encouraged to use ``2'' whenever they have doubts if a sample contains a cyberbullying
or not. Splitting the whole annotation process into two stages was considered since the beginning. The goal of the first
stage was to split the dataset into equivocal and unequivocal samples. Then, the equivocal samples were annotated by
additional 3 annotators that have never seen the samples before. Based on these two stages, the final labels were
determined using all annotations and weighting schemes described in the next section.

The labeling criteria were defined by describing online violence phenomena in relation to the target of the violence.
Therefore, the main concerns were:
\begin{itemize}
  \item Which types of phenomena can be considered as cyberbullying (e.g. personal attacks, threats, blackmails)?
  \item Who must be the target of online violence behavior in order to consider the sample as cyberbullying (e.g.
    interlocutor, individuals or groups identified by names)?
\end{itemize}

As the task is to detect cyberbullying, not profanity or abusive language in general, the guidelines recommended to turn
a blind eye to any usage of bad language in other situations than abusing or offending other person or things that are
(or might be) important to this person.

\subsection{Inter-annotator Agreements and Weighting Scheme}
\label{sec:agreement}

Firstly, we looked at each particular annotator, to specify the order of their proficiency in data labelling. This would
allow us later to properly weight the annotators in case of unclear annotation results, and thus disambiguate the cases
for which the final decision whether a sample (sentence, Internet entry) was harmful or not couldn't be easily made.

After the annotation process was complete, we calculated how the annotaters agreed between one another. As a mean for
ranking the annotators and verifying inter-annotator agreement we decided to use a simple percentage of the same
agreements within all applicable annotations, which was calculated for each pair of annotators. The overall annotator
ranking score was calculated as an average of all agreements with additional information provided by standard deviation.

The reason for not using standard \textit{kappa} \cite{fleiss1973equivalence} for calculating inter-annotator agreement
was as follows. Typical kappa is calculated for two classes, while in our class we allowed the third class ("uncertain''
/ ``I don't know"). Moreover, kappa assumes that there is an ideal answer (such as a type of disease to detect), whereas
for cyberbullying the case is more complicated. Although there are strict rules which define that something is
potentially a cyberbullying Internet entry, the ultimate decision whether something is or is not a cyberbullying is how
the bullied person feels about it.  Therefore, unless a first-person perspective evaluation is possible (which is
difficult if not impossible to obtain in practice), the decision has to be made based on the expert knowledge of the
annotators, with the initial assumption that all of them are equally capable to perform the task.

\begin{figure}
  \begin{center}
    \includegraphics[width=0.43\textwidth]{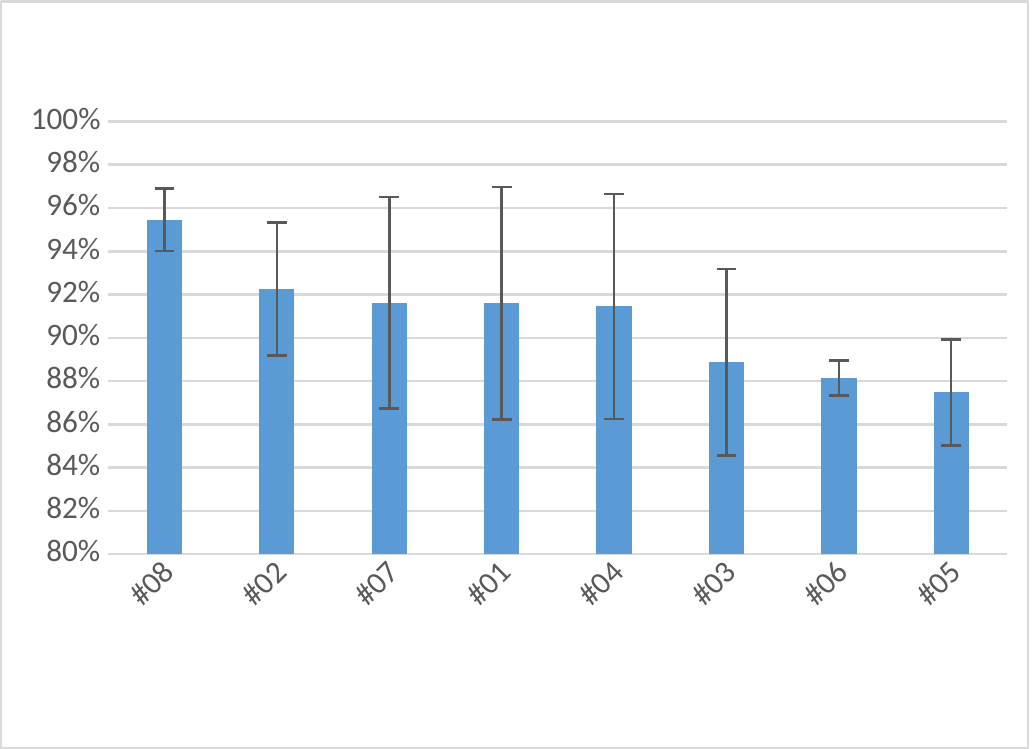}
  \end{center}
  \caption{Agreements between annotators in the first stage of annotation.}
  \label{fig:agreement1}
\end{figure}

There was no correlation (based on Pearson Rank Correlation coefficient) between annotator score and initial test
results. The Table \ref{tab:agreement1} and Figure \ref{fig:agreement1} give the mean agreements (with standard
deviation) for each annotator.

\begin{table}
  \begin{center}
    \caption{Agreements between annotators in the first stage of annotation.}
    \label{tab:agreement1}
    \begin{tabular}{|r|r|r|}
      \hline
      \multicolumn{1}{|r|}{\textbf{ID}} &
      \textbf{Mean agreement} &
      \textbf{Standard deviation} \\
      \hline
      \#08 &
      95.4\% &
      1.4\% \\
      \hline
      \#02 &
      92.2\% &
      3.1\% \\
      \hline
      \#07 &
      91.6\% &
      4.9\% \\
      \hline
      \#01 &
      91.6\% &
      5.4\% \\
      \hline
      \#04 &
      91.4\% &
      5.2\% \\
      \hline
      \#03 &
      88.9\% &
      4.3\% \\
      \hline
      \#06 &
      88.1\% &
      0.8\% \\
      \hline
      \#05 &
      87.5\% &
      2.4\% \\
      \hline
    \end{tabular}
  \end{center}
\end{table}

The average of all overall agreement scores among the annotators was 90.86\%. Three of the annotators were below the
average, and we considered them as the ``low performance group''. However, since the data was annotated each time by three
different annotators, we did not assume a top down threshold for weighting the annotators, but verified the weighted
rank for each sample based on each annotator's average strength of agreement.

\subsection{Data Annotation Results}

\subsubsection{I Stage of Annotation}
After the analysis of annotator performance, we looked at the data from the point of view of the annotated samples.
Therefore, we focused on agreements calculated per sample, and not per annotator. Most of the samples (almost 90\%) were
annotated unequivocally, which was a positive result suggesting good quality of the dataset and supporting the good
performance of annotators. Grand majority (86\%) of annotations was considered as \textit{not harmful}. Only about 2\% was
considered \textit{harmful} without a doubt by all annotators in the first stage of annotations. There was also a small number
(28) of samples for which none of the annotators had any idea how to classify them (\textit{uncertain} / \textit{I don't
know}), and also a small number of samples (17) which were not annotated, probably by accident. The 28 samples which all
three annotators were not able to annotate as well as the missing annotations were forwarded to a cyberbullying expert
for final decision. The remaining 1452 samples (11.37\%) annotated equivocally with some level of disagreement were used
in the second stage of annotation. Table \ref{tab:annotationR} contains the general outline of the dataset after the
first stage of annotations.

\begin{table}
  \begin{center}
    \caption{The outline of the dataset after the first stage of annotation.}
    \label{tab:annotationR}
    \begin{tabular}{|l|r|r|}
      \hline
      \textbf{Type} &
      \textbf{Number} &
      \textbf{Percentagen} \\
      \hline
      non-harmful &
      11007 &
      86.18\% \\
      \hline
      harmful &
      285 &
      2.23\% \\
      \hline
      I don't know &
      28 &
      0.22\% \\
      \hline
      equivocal &
      1452 &
      11.37\% \\
      \hline
      \textbf{sum} &
      \textbf{12772} &
      100.00\% \\
      \hline
    \end{tabular}
  \end{center}
\end{table}

Since most of the samples were annotated unequivocally, in the later stage of data preparation we focused only on those
samples, which were problematic (ambiguous, or annotated equivocally). To eliminate, and further specify the equivocal
annotations we performed two types of analysis. Firstly, we performed the second stage of annotations, as described in
Section \ref{sec:agreement}. However, relying only on the new second stage annotations and not taking into
account also the first annotations whatsoever could introduce an additional bias. It could also add even more ambiguity
to the samples which were only slightly ambiguous. Therefore, we also created an analysis procedure to weight all
ambiguous annotations depending on how ambiguous they were. Finally, we compared both: weighted first stage ambiguous
annotations and second stage annotations (also weighted for ambiguous cases). The analysis was done to eliminate the
ambiguity to some extent, or to somehow estimate the annotation class value even if the annotation did not reveal a
clear cut. Finally, after the comparison of two annotation attempts, all the annotations that were left, which could not
be disambiguated, were forwarded to the cyberbullying expert to obtain final verdict.

\subsubsection{Disambiguation Procedure for Equivocal Samples}

Since each data sample was annotated by three annotators, we divided the equivocal samples as follows.

\begin{enumerate}
  \item if one \textit{I don't know} case $\rightarrow$ remaining two decide
    \begin{enumerate}
      \item if remaining two were unequivocal $\rightarrow$ OK (disambiguation complete)
      \item if remaining two were scrambled $\rightarrow$ applying inter-annotator agreement-based weighting scheme +
        expert check for final decision
    \end{enumerate}
  \item if two \textit{I don't know} cases $\rightarrow$ weighting scheme + expert check for final decision
  \item if no \textit{I don't know} cases, but results scrambled (001, or 110) $\rightarrow$ weighting scheme + expert check
\end{enumerate}

There were only four possible situations of ambiguity. Firstly, if one of the annotators selected \textit{I don't know} (later
\textit{IDK} or \textit{uncertain}), or there was no annotation, the remaining two annotators were taken into consideration. Here, if
both selected the same answer, we considered the case as solved (however, we still checked those samples in the second
stage of annotations for final confirmation). If the results were scrambled, we applied appropriate weighting scheme to
propose an initial proposed estimated value (either 1 for \textit{harmful} or 0 for \textit{non-harmful}), and asked an additional
cyberbullying expert to verify the choice. Weighting was based on each annotator's average agreement score (the higher
the better).  Also, when none of the annotators selected \textit{IDK} but the results were still  scrambled, we also
applied an appropriate weighting scheme with expert's verification. Finally, when there were two \textit{IDK} answers,
we considered the remaining non ambiguous answer as a potentially correct, but with a strong voice from the second stage
of annotations and the additional expert.
Table \ref{tab:scrambled} contains the outline of the distribution of samples.

\begin{table}
  \begin{center}
    \caption{The outline of the ambiguous results and the disambiguation scheme for each type of ambiguity.}
    \label{tab:scrambled}
    \begin{tabular}{|l|r|l|}
      \hline
      \textbf{Type of Ambiguity} &
      \textbf{No. of Cases} &
      \textbf{Disambiguation} \\
      \hline
      has ``I don't know'' &
      1176 &
      \\
      \hline
      $\hookrightarrow$ 1 IDK (all) &
      931 &
      OK \\
      \hline
      \hspace{0.25cm} $\hookrightarrow$ 1 IDK unequivocal &
      762 &
      weighting + expert \\
      \hline
      \hspace{0.25cm} $\hookrightarrow$ 1 IDK scrambled &
      169 &
      expert \\
      \hline
      $\hookrightarrow$ 2 IDK &
      245 &
      weighting + expert \\
      \hline
      scrambled &
      276 &
      weighting + expert \\
      \hline
      \textbf{SUM} &
      1452 &
      \\
      \hline
    \end{tabular}
  \end{center}
\end{table}

Apart from weighting the annotators, we assigned a custom score of annotation confidence to all annotations. The
confidence score could be either \textit{high}, \textit{medium}, or \textit{low}, and was assigned according to the following principles.

Firstly, all non-ambiguous annotations (all three agreements) were considered to have a high confidence. The annotations
for which one of the answers was \textit{IDK}, but the other two were unequivocal, were also considered as \textit{high}. Samples which
had scrambled annotations were assigned \textit{high} confidence if two top-weight annotators agreed, \textit{medium} if first and
last agreed, and \textit{low} if only two low-weight annotators agreed.

All other annotations were assigned \textit{low} confidence. This accounts for the following situations. The samples with two
IDK annotations represent a situation, where two of the annotators were not sure what class to assign to the sample.
Therefore, even if the third annotator proposed a not-\textit{IDK} annotation, it cannot be considered as sufficiently certain,
and should be checked by an additional expert. Also, if one annotator did not know what to assign, and the other two
disagreed, this too represents a situation, where a clear answer cannot be drawn only from the three annotators and
should be checked by additional expert.

It also has to be added that whether an annotator was considered as high-weight or low-weight annotator was not
specified top-down only on the basis of the annotator's average agreement score, but rather calculated separately for
each sample. Since each sample was annotated by different set of annotators, the three annotators for each sample were
considered to have high, medium or low weight depending on how their average agreement related to other two annotators
for the sample.

\subsubsection{II Stage of Annotations and Final Disambiguation of Samples}

\begin{figure}
  \begin{center}
    \includegraphics[width=0.43\textwidth]{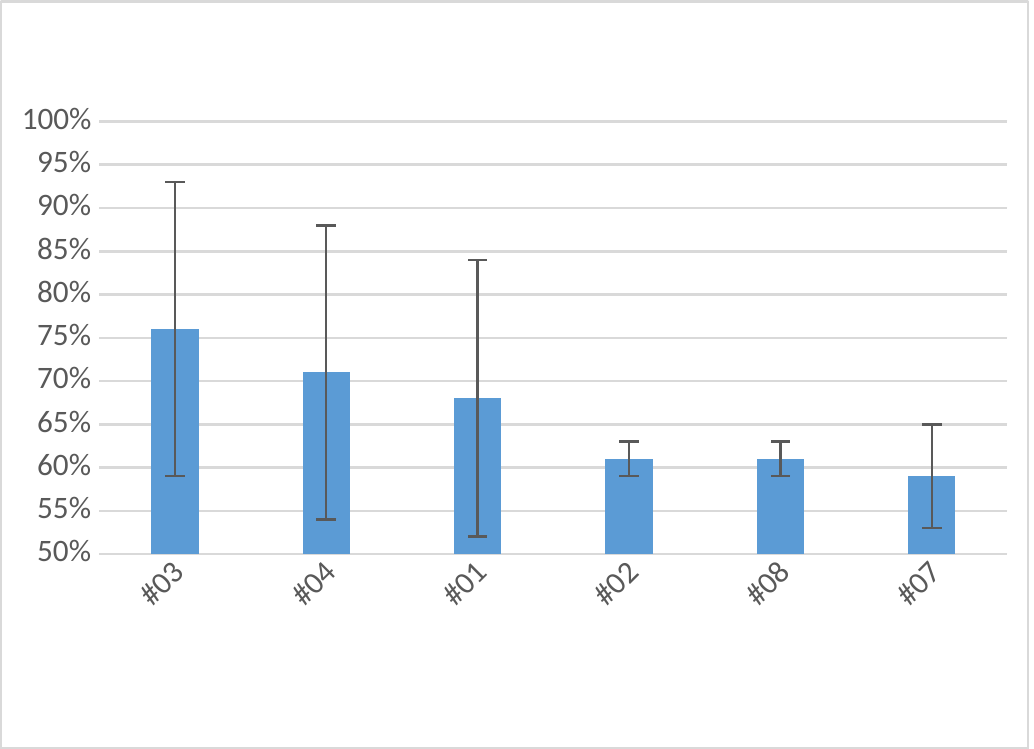}
  \end{center}
  \caption{Agreements between annotators in the second stage of annotation.}
  \label{fig:agreement2}
\end{figure}

In the second stage of annotations we applied only six annotators who decided to perform the additional annotations. The
samples were assigned to the annotators randomly, with a strict rule that the annotators did not see or annotate the
samples in the first step of annotations. Even if the annotator obtained a high agreement previously, their annotation
performance could deteriorate over time. Therefore, the mean agreements for the second stage were also calculated and
considered separately from the first stage. Overall average agreements were in general much lower for the second stage
of annotations. This confirms that the cases that were annotated were in general more problematic to annotate, and the
disambiguation did not result from annotators' personal performance. Table \ref{tab:agreement2} and Figure
\ref{fig:agreement2} contain the overall average agreements of the annotators that took part in the second stage of
annotations.

\begin{table}
  \begin{center}
    \caption{Agreements between annotators in the second stage of annotation.}
    \label{tab:agreement2}
    \begin{tabular}{|r|r|r|}
      \hline
      \multicolumn{1}{|r|}{\textbf{ID}} &
      \textbf{Mean agreement} &
      \textbf{Standard deviation} \\
      \hline
      \#03 &
      76\% &
      17\% \\
      \hline
      \#04 &
      71\% &
      17\% \\
      \hline
      \#01 &
      68\% &
      16\% \\
      \hline
      \#02 &
      61\% &
      2\% \\
      \hline
      \#08 &
      61\% &
      2\% \\
      \hline
      \#07 &
      59\% &
      6\% \\
      \hline
    \end{tabular}
  \end{center}
\end{table}

\begin{table}
  \begin{center}
    \caption{The outline of the dataset after the second stage of annotation.}
    \label{tab:scrambled2}
    \begin{tabular}{|l|r|l|}
      \hline
      \textbf{Type of Ambiguity} &
      \textbf{Number} &
      \textbf{Percentage} \\
      \hline
      All Ambiguous &
      1452 &
      \\
      \hline
      2$^{nd}$ annotation = unequivocal &
      753 &
      52\% (of all amb.) \\
      \hline
      $\hookrightarrow$ confirmed = SOLVED &
      652 &
      87\% (of unequiv.) \\
      \hline
      $\hookrightarrow$ unconfirmed = NEED CHECK &
      101 &
      13\% (of unequiv.)\\
      \hline
      2$^{nd}$ annotation = equivocal &
      699 &
      48\% (of all amb.)\\
      \hline
    \end{tabular}
  \end{center}
\end{table}

The results of the second annotation are given in Table \ref{tab:scrambled2}. It solved over half of the problem of
ambiguity. For this part we compared the score with the first
annotations. If the score was the same, we considered it as the final score. This accounted for 87\% of unequivocal cases
(in the second stage),
which in most part confirmed most of the first stage annotations, thus reassuring us about the quality of annotations.
Next, if the score was different (13\% of cases), the expert performed confirmation check to make the final decision on
which result was correct. Out of the remaining half (48\%) of the equivocally annotated cases, we performed the initial
automatic disambiguation based on average agreement of each annotator, compared the results to the first stage
annotations and had the cyberbullying expert double check all the remaining cases to make the final decision.

The final annotated data after all disambiguation and final check by the expert contained 913 cyberbullying cases (7\%)
within the overall 12772 cases. The results are summarized in Table \ref{tab:final}.

\begin{table}
  \begin{center}
    \caption{Final results of the annotation after two stages of annotation and expert's decision for the hardest cases.}
    \label{tab:final}
    \begin{tabular}{|l|r|r|}
      \hline
      \textbf{Type} &
      \textbf{Number} &
      \textbf{Percentage} \\
      \hline
      harmful &
      913 &
      7\% \\
      \hline
      non-harmful &
      11859 &
      93\% \\
      \hline
      \textbf{total} &
      \textbf{12772} &
       \\
      \hline
    \end{tabular}
  \end{center}
\end{table}

Our final annotations contained more cyberbullying cases than the original annotations performed by previous
researchers. We were also able to both correct the cases wrongly annotated as \textit{harmful}, and those incorrectly annotated
as \textit{non-harmful} (mentioned in Section \ref{sec:limitations}). Since the quality of annotations
was increased, we also assumed that this would improve the objectivity of the evaluation of the proposed system, as well
as help develop more accurate machine learning model in the final hybrid system.

\subsection{Dataset Properties}
\label{sec:stats}

Table \ref{tab:stats} reports some key statistics of the dataset. Statistics described as \textit{harmful} and
\textit{non-harmful} refer to the final version of the new annotations. The dataset contains approximately 300 thousand of
tokens, making it rather small regarding current ML trends. There are no big differences in length between the
posted questions and answers (approx. 12 words). On the other hand, the \textit{harmful} samples are usually a bit shorter than
the \textit{non-harmful} samples (approx. 23 vs. 25 words). The number of \textit{harmful} examples is small, amounting to only 7\%, yet
it seems to be rather high compared to the average content of social networks.

\begin{table}
  \begin{center}
    \caption{Statistics of the dataset computed after final annotation of the data.}
    \label{tab:stats}
    \begin{tabular}{|l|r|}
      \hline
      \textbf{Element type} & \textbf{Value} \\
      \hline
      Number of samples &
      12772 \\
      \hline
      Number of \textit{harmful} samples  &
      913 \\
      \hline
      Number of \textit{non-harmful} samples &
      11859 \\
      \hline
      Number of all tokens &
      301198 \\
      \hline
      Number of unique tokens &
      18394 \\
      \hline
      \hline
      Avg. length (characters) of a single post (Q + A) &
      120.1 \\
      \hline
      Avg. length (words) of a single post (Q + A) &
      23.6 \\
      \hline
      Avg. length (characters) of a single question &
      61.6 \\
      \hline
      Avg. length (words) of a single question &
      12.0 \\
      \hline
      Avg. length (characters) of a single answer &
      58.5 \\
      \hline
      Avg. length (words) of a single answer &
      11.5 \\
      \hline
      Avg. length (characters) of a \textit{harmful} post &
      120.1 \\
      \hline
      Avg. length (words) of a \textit{harmful} post  &
      22.9 \\
      \hline
      Avg. length (characters) of a \textit{non-harmful} post &
      130.9 \\
      \hline
      Avg. length (words) of a \textit{non-harmful} post &
      24.7 \\
      \hline
    \end{tabular}
  \end{center}
\end{table}

\subsection{Comparison with the Previous Annotation}

The original annotation provides 802 samples out of 12772 (6.3\%) labeled as \textit{cyberbullying} (or \textit{harmful}), according to
the proposed method of classifying a post as \textit{cyberbullying} if it was labeled as \textit{cyberbullying} by at least 2 out of 3
annotators \cite{reynolds2011using}. The new annotation provides 913 samples out of 12772 (7.1\%) labeled as \textit{harmful.}

There are 392 samples that were labeled as \textit{non-harmful} in the original annotation and as \textit{harmful} in the new
annotation. 9 out of 10 examples from Section \ref{sec:limitations} were correctly labeled as
\textit{harmful} in the new annotation. Some additional examples are given in Appendix B%
.

There are 281 samples that were labeled as \textit{harmful} in the original annotation and as \textit{non-harmful} in the new
annotation. 10 out of 10 examples from section \ref{sec:limitations} were correctly labeled as
\textit{non-harmful} in the new annotation. Some additional examples are given in Appendix B%
.

\section{Samurai - Cyberbullying Detection System}

In this section we present Samurai -- the cyberbullying detection system, its simplified architectural workflow, its key
features compared to the other state-of-the-art approaches to the cyberbullying detection, the onboarding process, and
the way the system was provided for testing.

\subsection{System Overview}

Samurai is a hybrid AI prototype system for detecting cyberbullying. It utilizes statistical components (e.g. deep learning
modules) under the strict government of symbolic modules. A statistical component can be used to perform certain
well-defined sub-tasks (e.g. phrase classification), but there is always a symbolic module that is responsible for
making the final decision whether or not a text contains cyberbullying. This approach enables the decision-making
process to be largely explainable and trackable.

A few examples of combining statistical components with symbolic governance within the system:
\begin{enumerate}
  \item A statistical component is used to determine if a given expression can be considered as abusive, whereas a
    symbolic module is responsible for determining if the expression is targeted against an interlocutor (e.g. by the
    use of a linking verb).
  \item A symbolic module is used to detect all conditional constructions as potential candidates, and to split them
    into condition and consequence parts. Then, a statistical component is used to evaluate harmfulness of the
    consequence part, whereas another symbolic part is responsible for verifying -- based on the statistical evaluation
    -- if the whole candidate can be considered as a blackmail or not.
  \item The highest-level counterfactual verification of the detected cases that is able to determine that the sentence
    ``You are an idiot.'' is a personal attack, whereas ``I don't think you are an idiot.'' is not.

\end{enumerate}

The symbolic modules strongly rely on the grammar structure of a processed text. The grammar structure is given through
a deep syntactic analysis provided by a dedicated syntactic parser – Language Decoder (a proprietary and patented
technology of Samurai Labs). The methodology of building precise models based on grammar structure involves elements of
Context-based Information Extraction (another proprietary and patent-pending technology of Samurai Labs).

Samurai utilizes a multi-level modular architecture, where the top-level division is comprised of a text preparation
engine and a detection engine. The detection engine consists of a set of dedicated modules, where each module is
responsible for detecting a specific online violence phenomenon. Each module is comprised of a set of sub-modules
specialized in finding certain aspects of the given phenomenon. For example, a sub-module detecting abusive comparisons
towards an interlocutor is a part of a module responsible for personal attacks detection. Under each sub-module there
are groups of dedicated rules that directly set logical constraints on the grammar structure. As a result, Samurai is
able to provide a multi-level hierarchical categorization of the detected phenomena and their aspects.

The detection engine is preceded by the text preparation engine that contains normalization, correction and
transformation modules. Its purpose is to prepare input text for the detection process. Normalization is the process of
adjusting an input text into a predefined standard format (e.g. support for encoding, emoticons, special characters and
segmentation). Correction is the process of revising the normalized text in order to remove misspellings and any other
errors that may impede the work of the detection engine. It covers correction-related tasks from handling abbreviations
and typos to solving the complex problems with grammatical errors based on the context. The correction module is also
responsible for detecting attempts to cheat the system such as using various spelling combinations (e.g. switching
letters with numbers or similarly looking unicode characters). Transformation performs operations on normalized and
corrected text that makes it more suitable for the detection engine, including support for idiomatic expressions, some
aspects of coreference resolution and filling a sentence with omitted words or phrases (e.g. pronouns or infinitive
particles).

\subsection{Onboarding Process}

The onboarding is a process that adjusts the system to the client's needs. It starts with the client defining what would
be the desired system's reaction to certain phenomena. The process of gathering this information can be performed in any
form of question and answer interviews (or surveys), including a simple online survey on the product's website. The
concept is that for every question the client decides ``yes / no'' whether the system should ``block'' or ``pass'' a given
phenomenon. Exemplary questions from the standard onboarding process are given in Appendix B%
.

Question can appear in several different levels of intensity -- from almost non-offensive to highly offensive. If the
answer for the current question is ``pass,'' then a more offensive version is presented. For example, a client wants the
non-abusive questions about sexuality to pass through. The next version of the question would be: what is the desired
reaction for coarse questions about sexuality. This hierarchical process allows the system to set the proper boundaries
to determine what should be blocked and what should pass through.

Once the questions are answered, a trained engineer adjusts the system based on the desired guidelines. Due to the
multi-level modular architecture, this process resembles switching on and off certain nodes on a decision tree. For
example, if a client wants the non-abusive questions about sexuality to pass through, an engineer goes to the sexual
harassment module in the detection engine, then to the submodule responsible for detecting questions about sexuality,
and finally to the section related to non-abusive topics of the questions. Only the latter one becomes disabled, which
fulfills the client's need without interrupting any other parts of the system. At the end, the customized version of the
API is released and the client is provided with the proper access.

For the purpose of this research, a person without significant engineering skills and linguistic knowledge was asked to
role-play the client for the onboarding process. The ``client'' was provided with the final version of the re-annotated
dataset. The questions were asked in batches so that the ``client'' had enough time to find the answers in the dataset.
One trained engineer was involved in the process of adjusting the system on the Samurai Labs' side. The whole onboarding
process lasted 3 workdays. The final annotations were delivered on June 12, 2018, the process started on June 13, 2018
and ended up on June 17, 2018. Aside from this standard onboarding process, the newly labeled dataset was never used in
the process of building or refining the Samurai system.

\subsection{System API}

For the purpose of preparing this report, the standard Samurai Labs' procedures were applied. The Samurai was provided
in a form of dedicated API after the onboarding process described in the previous section. The system is cloud-hosted.
The API takes in a single text as a POST request.

An exemplary API request as a cURL command:

\texttt{curl -H "x-api-key: <API KEY>" \\
\indent --data-urlencode "text=<TEXT TO PROCESS>" <API URL>}

After the onboarding process, the client is provided with the API KEY and the API URL, which have to be provided as parameters to the POST call.

The API responds with JSON containing a score representing an online violence level in a scale of 0 to 1, and a list of
detected online violence categories (e.g. direct abuse towards interlocutor). The online violence level is represented
with a score from 0 (no violence) to 1 (high level of violence), with a threshold at 0.7 (medium level of violence). The
API is set up to process 20 texts per second.

Few example API responses are given in Appendix B%
.

\section{Experiment}

In this section we present procedures and results of testing the Samurai system along with one well-known classification
algorithm (Fasttext) and five commercial products for cyberbullying detection and content moderation, that are currently
available on the market.  Additionally, for the Samurai system, an error analysis is
provided to further evaluate the system's performance and the new annotation's quality.

\subsection{Procedure}
To assess the Samurai system and to check if there are important differences between the annotations the following
experiments were conducted. The researchers from AGH University of Science and Technology were given access to the API.
In the first experiment the original Formspring dataset with its annotation was used. An entry was counted as
\textbf{harmful}, if at least two of the three annotators marked it as such. Therefore, an entry was counted as \textbf{not
harmful} if the majority of the annotators agreed that it is not harmful. For the system, an entry was considered
\textbf{harmful}, if it received score 0.7 or above. Thus \textit{true positive} was counted when both the annotation
and the system marked given sample as harmful, \textit{false positive} when the system considered it harmful, while the
annotation not and \textit{false negative} when the system considered a sample as not harmful, while the annotation as
harmful.

In the second experiment the new annotation was used. The \textbf{harmful} samples were established by the procedure described in
Section \ref{sec:dataset}. Besides that the \textit{true positives}, \textit{false positives} and \textit{false
negatives} were defined the same as in the first
experiment. After a first round of queries, it turned out, that some of the results were different between what was
reported by the creators of the system and the results obtained by AGH researchers. Since the researchers didn't have
access to the system, they inspected the samples that were causing the differences. It turned out that the they had one
thing in common: presence of HTML entities. The samples were processed to convert these HTML entities into regular
characters. After the conversion, there was no difference between the results reported by the creators of the system and
the AGH researchers. Regarding the performance of the system -- there were 2 timeouts. Repeating the calls for the same
samples caused another timeouts, thus it seems that the error is connected with these examples. The problem was reported
to the Samurai team.

To compare the performance of the system with an off-the-shell classification algorithm,  Fasttext \cite{joulin2017bag}
was selected as one of algorithms that performs particularly good in many NLP-related tasks. The system was run with the
following parameters:
\begin{verbatim}
supervised -dim 10 -lr 0.1 -wordNgrams 2 -minCount 1
  -bucket 10000000 -epoch 5 -thread 4
\end{verbatim}
The testing procedure for Fasttext followed the 10-fold cross validation scheme.

Additionally, 5 commercial systems providing cyberbullying detection and content moderation (denoted as A, B, C, D and
E) were tested using the same datasets. In each case (including Fasttext) the available parameters were tuned to obtain
the best possible results (F1 measure was used as the optimization criterion).

\subsection{Results}

\begin{figure}
  \begin{center}
    \includegraphics[width=0.49\textwidth]{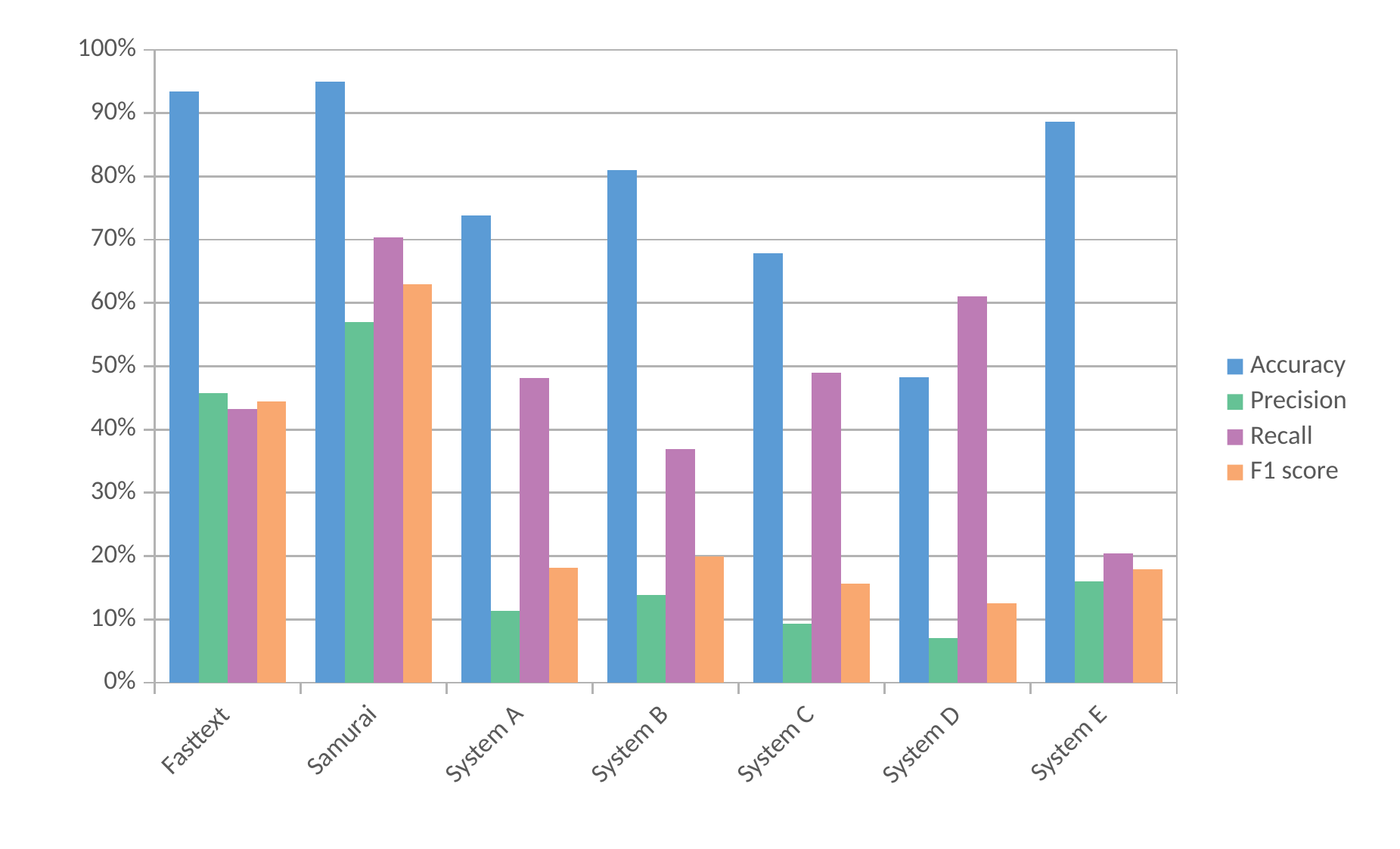}
  \end{center}
  \caption{Performance of the tested systems with respect to the \textit{old annotation}.}
  \label{fig:result1}
\end{figure}

\begin{figure}
  \begin{center}
    \includegraphics[width=0.49\textwidth]{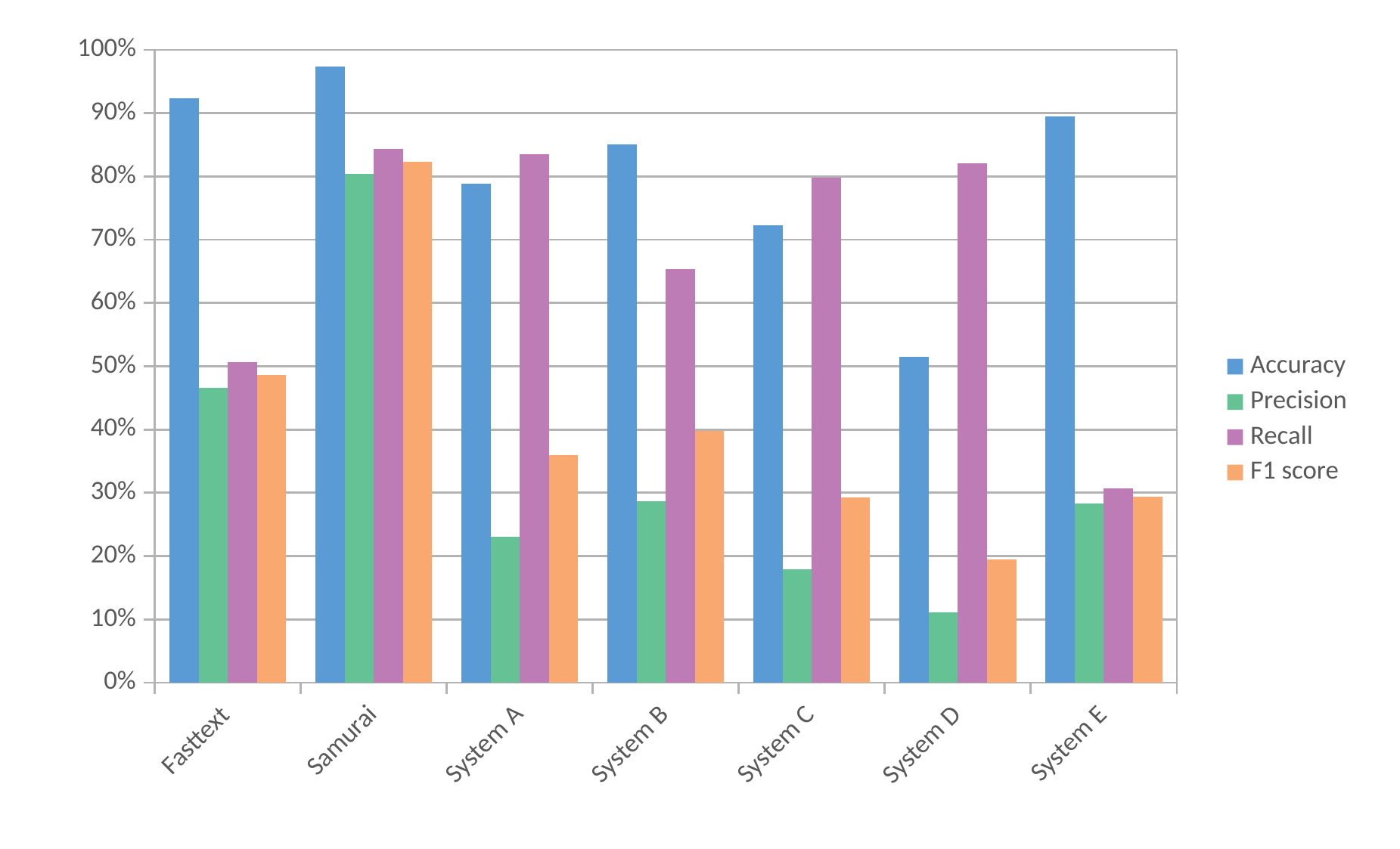}
  \end{center}
  \caption{Performance of the tested systems with respect to the \textit{new annotation}.}
  \label{fig:result2}
\end{figure}

The results of the experiment are summarized in Table \ref{tab:results}, Figure \ref{fig:result1} and Figure
\ref{fig:result2}. Accuracy, precision, recall and F1 are defined
in the standard way \cite{manning1999foundations}:

{\renewcommand{\arraystretch}{1.3}
\begin{tabular}{r c l}
  Accuracy &=& $\frac{TP + TN}{TP + TN + FN + FP}$ \\
  Precision &=& $\frac{TP}{TP + FP}$ \\
  Recall &=& $\frac{TP}{TP + FN}$ \\
  F1 &=& $\frac{2 * Pr * Rc}{Pr + Rc}$ \\
\end{tabular}} \\
In Table \ref{tab:raw} we also give raw values for \textit{true positives}, \textit{false positives}, \textit{true
negatives} and \textit{false negatives}, in order to emphasize the actual number of detected cases and false alarms, and
to stress the fact, that some of the examples caused API errors, which are not taken into account when computing the
above defined measures.

\begin{table}
  \begin{center}
    \caption{The results for cyberbullying detection for the tested systems computed using the
    \textit{old annotation} and the \textit{new annotation}.}
    \label{tab:results}
    \begin{tabular}{|l|r|r|r|r|}
      \hline
      \textbf{Algorithm} &
      \textbf{Accuracy} &
      \textbf{Precision} &
      \textbf{Recall} &
      \textbf{F1 score} \\
      \hline
      & \multicolumn{4}{|c|}{\textbf{Old annotation}} \\
      \hline
      Fasttext &
      0.934 &
      0.457 &
      0.432 &
      0.444 \\
      \hline
      Samurai &
      \textbf{0.950} &
      \textbf{0.570} &
      \textbf{0.704} &
      \textbf{0.630} \\
      \hline
      System A &
      0.738 &
      0.113 &
      0.481 &
      0.182 \\
      \hline
      System B &
      0.815 &
      0.132 &
      0.367 &
      0.194 \\
      \hline
      System C &
      0.679 &
      0.093 &
      0.490 &
      0.156 \\
      \hline
      System D &
      0.483 &
      0.070 &
      0.611 &
      0.125 \\
      \hline
      System E &
      0.886 &
      0.160 &
      0.204 &
      0.179 \\
      \hline
      & \multicolumn{4}{|c|}{\textbf{New annotation}} \\
      \hline
      Fasttext &
      0.923 &
      0.466 &
      0.507 &
      0.486 \\
      \hline
      Samurai &
      \textbf{0.974} &
      \textbf{0.804} &
      \textbf{0.843} &
      \textbf{0.823} \\
      \hline
      System A &
      0.788 &
      0.230 &
      0.835 &
      0.360 \\
      \hline
      System B &
      0.853  &
      0.277  &
      0.656  &
      0.390  \\
      \hline
      System C &
      0.723 &
      0.179 &
      0.798 &
      0.292 \\
      \hline
      System D &
      0.515 &
      0.111 &
      0.821 &
      0.195 \\
      \hline
      System E &
      0.895 &
      0.283 &
      0.307 &
      0.294 \\
      \hline
    \end{tabular}
  \end{center}
\end{table}

\begin{table}
  \begin{center}
    \caption{Raw results for cyberbullying detection for the tested systems computed using the
    the \textit{new annotation}. Differences in totals are due to errors in API calls.
    TP -- true positive, FP -- false positive, FN -- false negative, TN -- true negative.}
    \label{tab:raw}
    \begin{tabular}{|l|r|r|r|r|r|}
      \hline
      \textbf{Algorithm} &
      \textbf{TP} &
      \textbf{FP} &
      \textbf{TN} &
      \textbf{FN} &
      \textbf{Total} \\
      \hline
      Fasttext &
      463 &
      530 &
      11329 &
      450 &
      12772\\
      \hline
      Samurai &
      770 &
      188 &
      11669 &
      143 &
      12770
      \\
      \hline
      System A &
      762 &
      2553 &
      9284 &
      151 &
      12750
      \\
      \hline
      System B &
      599 &
      1562 &
      10297 &
      314 &
      12772\\
      \hline
      System C &
      729 &
      3355 &
      8503 &
      184 &
      12771\\
      \hline
      System D &
      750 &
      6028 &
      5831 &
      163 &
      12772
      \\
      \hline
      System E &
      280 &
      710 &
      11149 &
      633 &
      12772
      \\
      \hline
    \end{tabular}
  \end{center}
\end{table}

The Samurai system gives the best results in all of the measures on the new as well as the old annotation. Most
importantly, it gives the
best F1 score on the new annotation, which is significantly better than Fasttext (more than 80\% compared to 44\%-49\%
depending on the annotation). This difference is mainly due to high recall of the system, reaching nearly 85\%,
which is particularly important for the purpose of preventing the harmful aspects of cyberbullying. All the other
commercial systems perform worse than the Fasttext algorithm. Their low performance is due to
their low precision spanning from 7\% (system D on the old annotation) to 28.3\% (system E on the new annotation).

The results also show that the new annotation is more coherent than the original one since the machine learning
algorithm was able to give better precision and better recall on that dataset. Samurai and the other systems which were
not trained on the dataset also gave better results. The differences are pretty large, especially in terms of precision,
meaning that the new annotation is a much different dataset, than the old one.

\subsection{Error analysis}

Since no part of the Samurai system was trained or built using the newly labeled dataset (the system was adjusted using
the dataset in the onboarding process), it would be very informative to analyze the errors that the system made in
relation to the new annotations.

The error analysis was performed by a Samurai Labs' team member involved in the process of building the Samurai system.
For the sake of transparency, the whole analysis is publicly available in a read-only Google Sheets under
\url{https://goo.gl/frRiZP}.

The document is divided into two sections regarding false positives and false negatives, separately. Each error was
classified into one of three following categories (each represented by a separate column):
``CORRECT'': The annotators' decision is considered to be appropriate; the system is wrong;
``MAYBE'': The case is ``on the edge'' and the given criteria were not sufficient enough; the decision if it should be
blocked or passed through would require new criteria that take into consideration these phenomena;
``INCORRECT'': The annotators' decision is considered to be inappropriate (e.g. due to the arbitrary omission or clear
incompatibility with the annotation instruction / guidelines); it should not be considered as an error since the system
is right.

For the 188 false positives:
\begin{itemize}
  \item 44 (23.4\%) were considered as CORRECT,
  \item 79 (42\%) as INCORRECT, and
  \item 65 (34.6\%) were labeled as MAYBE.
\end{itemize}

Among false positives, the largest part comprises INCORRECT cases which draws a conclusion that even the high standards
of the annotation process allow some occurrences of cyberbullying to glide over.

A significant part of INCORRECT cases (16 out of 79) comprises posts containing phenomena that were labeled as ``harmful''
in other posts. As there is no significant discriminant between those two groups, they should be labeled in the same
manner. Otherwise, the annotation would be inconsistent. The other part (11 out of 79) comprises labels incompatible
with the annotation instruction / guidelines (point 2. in Online Violence Target section). The largest part (30 out of
79) comprises posts containing direct abuses towards an interlocutor, without a clear impression that the usage is
consensual. Among these cases, 12 of 30 are assignments that assign an abusive phrase to the interlocutor using linking
verbs, and 18 of 30 are abusive vocative cases.

The largest part of MAYBE cases (33 out of 65) also comprises posts containing direct abuses towards an interlocutor,
but these cases seem to be consensual based on the given context. Among these cases, 10 of 33 are assignments that
assign an abusive phrase to the interlocutor using linking verbs, and 23 of 33 are abusive vocative cases. A significant
part (8 out of 65) comprises posts containing mild sex-related content. Although they are not abusive towards an
interlocutor, the decision whether they should be ``blocked'' or ``passed through'' should be unambiguously defined by the
proper guidelines. Most of the other cases can also be considered as abusive towards an interlocutor but with a clear
impression that interlocutors jointly agree for the given form of the conversation.

For the 143 false negatives:
\begin{itemize}
  \item 68 (47.5\%) were considered as CORRECT,
  \item 25 (17.5\%) as INCORRECT, and
  \item 50 (35\%) were labeled as MAYBE.
\end{itemize}

Among false negatives, the largest part comprises CORRECT cases which is an expected situation. Although, both MAYBE and
INCORRECT cases cover more than half of all false negatives and therefore require at least brief elaboration.

A significant part of MAYBE cases (25 out of 50) comprises posts containing mild sex-related content that is not abusive
towards an interlocutor or any other person. The next significant part (10 of 50) comprises posts contains coarse
language (including 4 examples of sex-related content) that remain not abusive towards an interlocutor.

The INCORRECT cases comprise posts that do not contain sex-related content and it is very hard to consider them as
cyberbullying even under a very rigorous criteria. Most of them resemble normal conversation between two people that
jointly agree for the given form of the conversation. Some of them (5 out of 25) contain coarse language, but it is
clearly not used to offend an interlocutor.

The future work with the dataset should take into consideration all these cases, especially due to the fact that they
can be grouped into well defined categories. The analysis of false positives shows that despite the high standards of
the annotation process, some significant number of cyberbullying cases was still able to sneak through, including some
violences of the annotation guidelines. The analysis of false negatives shows that especially the criteria telling what
to do with mild sex-related content should be defined unambiguously and with a great attention before the annotation
process. Three methods are proposed to improve the annotation process:
\begin{enumerate}
  \item To prepare more precise annotation instruction / guidelines with a number of illustrative (positive and
    negative) examples, especially ``on the edge'' examples.
  \item To ensure that annotators work in small batches as tiredness is one of the key causes for making mistakes such
    as violence of the annotation guidelines.
  \item To set some checkpoints during the process when annotators can freely discuss their thoughts and doubts among
    themselves and with external experts.
\end{enumerate}

\section{Conclusions}

The in-depth analysis of the Formspring dataset performed during the annotation process showed that the original
annotation was not perfect. Although, in the case of any NLP task it is hard to say that any annotation is
\textit{perfect}, the results of evaluation of many cyberbullying detection algorithms as well as the results of
training a machine learning algorithm indicate that the new annotation is more coherent. We expect that it might become
a new reference annotation for this task.

Samurai demonstrates that its novel approach can comprise an effective way for cyberbullying detection. In that case,
high recall goes side by side with high precision, which indicates the possibility of using the system for real-time
automatic cyberbullying blocking and
content moderation. Furthermore, both the annotation process and the error analysis show how much depends on
the adopted criteria, and therefore how important it is for a cyberbullying detection system to be effectively
adjustable to the given criteria.

\bibliographystyle{IEEEtran}
    \bibliography{fido_report}

\end{document}